\UseRawInputEncoding
\documentclass[conference]{IEEEtran}
\IEEEoverridecommandlockouts

\usepackage{tikz}
\usepackage{textcomp}
\usepackage{hyperref}
\usepackage{lipsum}

\newcommand\copyrighttext{%
  \footnotesize \textcopyright 2021 IEEE. Personal use of this material is permitted.
  Permission from IEEE must be obtained for all other uses, in any current or future 
  media, including reprinting/republishing this material for advertising or promotional 
  purposes, creating new collective works, for resale or redistribution to servers or 
  lists, or reuse of any copyrighted component of this work in other works.  
  }
\newcommand\copyrightnotice{%
\begin{tikzpicture}[remember picture,overlay]
\node[anchor=south,yshift=10pt] at (current page.south) {\fbox{\parbox{\dimexpr\textwidth-\fboxsep-\fboxrule\relax}{\copyrighttext}}};
\end{tikzpicture}%
}

\usepackage{cite}
\usepackage{amsmath,amssymb,amsfonts}
\usepackage{algorithmic}
\usepackage{graphicx}
\usepackage{textcomp}
\usepackage{xcolor}
\usepackage{mdframed}
\usepackage{subfig}
\usepackage{float}
\def\BibTeX{{\rm B\kern-.05em{\sc i\kern-.025em b}\kern-.08em
    T\kern-.1667em\lower.7ex\hbox{E}\kern-.125emX}}

\begin{document}

\title{ViSTA: a Framework for Virtual Scenario-based Testing of Autonomous Vehicles

}

\author{\IEEEauthorblockN{ Andrea Piazzoni}
\IEEEauthorblockA{\textit{IGP-ERI@N and CETRAN} \\
\textit{Nanyang Technological University}\\
Singapore \\
andrea006@ntu.edu.sg}
\and
\IEEEauthorblockN{ Jim Cherian}
\IEEEauthorblockA{\textit{CETRAN} \\
\textit{Nanyang Technological University}\\
Singapore \\
jcherian@ntu.edu.sg}
\and
\IEEEauthorblockN{ Mohamed Azhar}
\IEEEauthorblockA{\textit{CETRAN} \\
\textit{Nanyang Technological University}\\
Singapore \\
mohd.azhar@ntu.edu.sg}
\and
\IEEEauthorblockN{Jing Yew Yap}
\IEEEauthorblockA{\textit{CETRAN} \\
\textit{Nanyang Technological University}\\
Singapore \\
jingyew.yap@ntu.edu.sg}
\and
\IEEEauthorblockN{James Lee Wei Shung}
\IEEEauthorblockA{\textit{CETRAN} \\
\textit{Nanyang Technological University}\\
Singapore \\
james.leews@ntu.edu.sg}
\and
\IEEEauthorblockN{Roshan Vijay}
\IEEEauthorblockA{\textit{CETRAN} \\
\textit{Nanyang Technological University}\\
Singapore \\
rvijay@ntu.edu.sg}
\thanks{This work was supported by the Centre of Excellence for Testing \& Research of AVs - NTU (CETRAN), Singapore. https://cetran.sg}
}

\maketitle
\copyrightnotice
\begin{abstract}
In this paper, we present ViSTA, a framework for Virtual Scenario-based Testing of Autonomous Vehicles (AV), developed as part of the 2021  IEEE  Autonomous Test Driving AI Test Challenge. Scenario-based virtual testing aims to construct specific challenges posed for the AV to overcome, albeit in virtual test environments that may not necessarily resemble the real world.
This approach is aimed at identifying specific issues that arise safety concerns before an actual deployment of the AV on the road. In this paper, we describe a comprehensive test case generation approach that facilitates the design of special-purpose scenarios with meaningful parameters to form test cases, both in automated and manual ways, leveraging the strength and weaknesses of either. 
Furthermore, we describe how to automate the execution of test cases, and analyze the performance of the AV under these test cases.
\end{abstract}

\begin{IEEEkeywords}
 Autonomous Vehicles, Simulation, Virtual Testing, Scenarios
\end{IEEEkeywords}

\section{Introduction}
Virtual Testing is one of the crucial steps in the assessment of performance and safety of Autonomous Vehicles (AV).
The core component in this step is a high-fidelity virtual simulator.
These simulators provide a virtual world in which a virtual vehicle can navigate, including the physics engine that model its dynamics.
Given this setup, if this vehicle is controlled by the same Automated Driving System (ADS) software \cite{standard2021j3016} as would be operating on an actual AV, it is possible to make an in-depth study of how the AV would react to specific traffic scenarios.
However, the design of the traffic scenarios is a complex, tedious, and non-standard procedure.
In particular, traffic scenarios can be compared to a scene where the actors (other vehicles or pedestrians) perform their part, and the ego vehicle (the system under test) has to react accordingly.
For example, a simple scenario could involve a pedestrian that, upon the approach of the AV, starts to cross the road directly into its path. This will require a timely reaction from the AV; otherwise, there will be either a collision or a near miss, both outcomes being undesirable.
In this context, it is trivial to observe that the validity of the whole testing procedure is only as good as the test cases used.
Furthermore, the implementation of a designed test case can be hindered by the capability and constraints of tools available, which leads to additional limitations in the overall virtual testing.

This paper stems from the 2021 IEEE Autonomous Test Driving AI Test Challenge \cite{avchallenge}, hence, 
the focus of this study is to conduct virtual testing of Baidu Apollo \cite{Apollo} using SVL \cite{svl} simulation platform.
In order to achieve this, we have developed ViSTA, a framework and the corresponding tools to facilitate the design of scenario testing, scenario execution and analysis.
Our contributions can be summarized as follows:
\begin{itemize}
\item Developed a scenario-generation framework, that facilitates the design of tests with varying degrees of automation, ranging from random to manual (controlled) tests.
\item Developed an automated Scenario-execution framework, built on top of the default SVL Python interface, with additional capabilities for varying dynamics of the actors.
 \item Generated test cases using our framework using both existing scenario databases and newly crafted scenarios, complemented by expert knowledge and analysis of the Operational Domain (OD) and Operational Design Domain (ODD) \cite{standard2021j3016} assuming SAE L4+ automation level.
\item Prepared the final test suite with both manual and auto-generated test cases (selection).
\item Executed the final test suite through a  comprehensive evaluation process, that combines both objective and subjective judgement, leading to the identification of several issues and specific limitations in the ADS under test, i.e., Baidu Apollo.
\end{itemize}

\section{Related Works}
\begin{figure*}[t]
     \centering
     \includegraphics[trim=0.5cm 5cm 0.5cm 3.5cm,width=\textwidth]{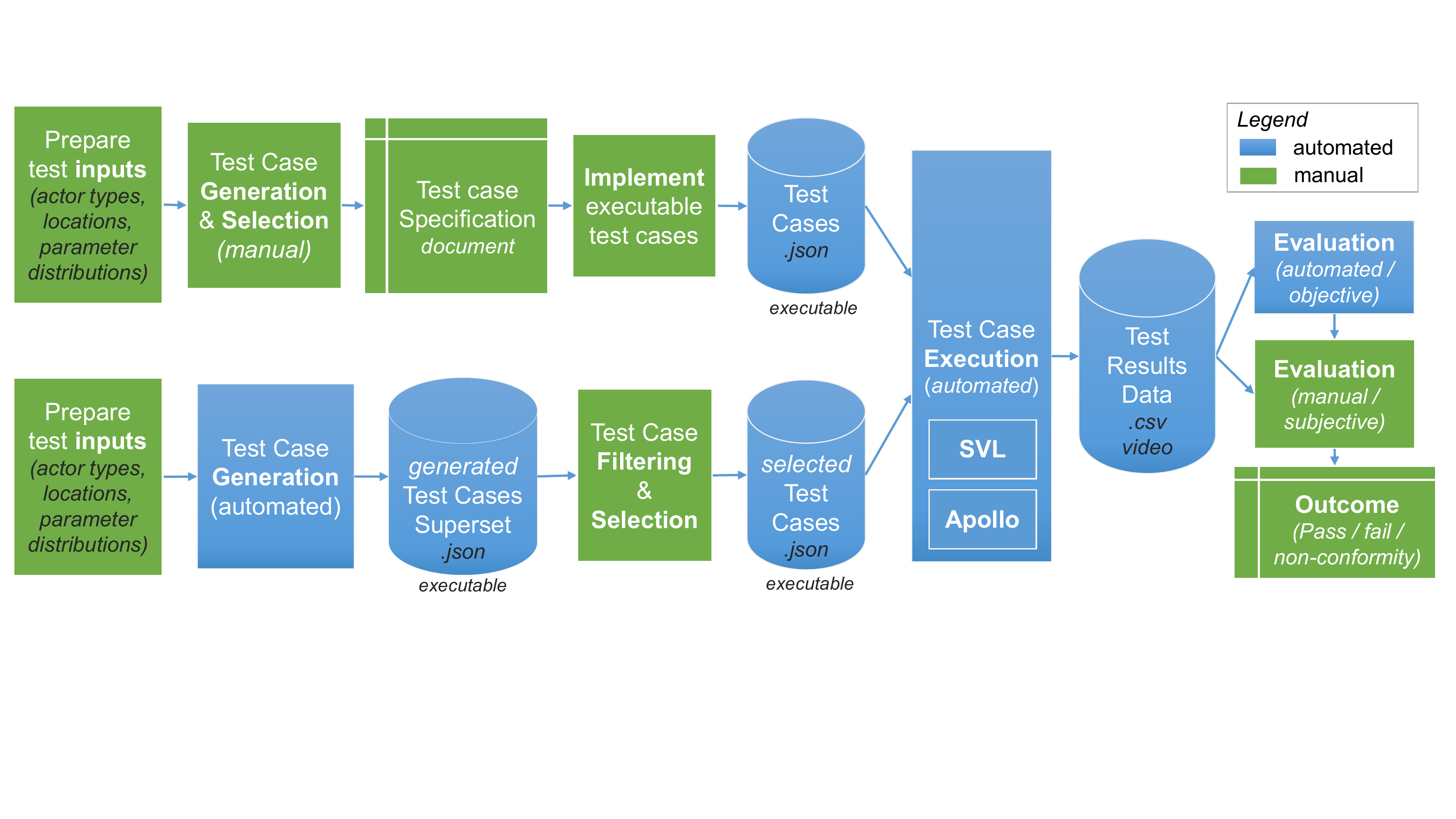}
     \caption{Overview of ViSTA framework: inputs, automated/manual virtual test generation, execution, evaluation and outcome}
     \label{fig:pipeline}
\end{figure*}
The development and testing of AI-based algorithms often relies on testing in equivalent virtual environments before exposing them to their actual physical deployment environments, especially in case of systems that require physical actuation, such as robotic systems.
To this end, in recent years, tools such as \cite{Brockman2016,Juliani2018} has become popular as they provide generic virtual environments, allowing researchers to focus on developing and testing the AI algorithms.
Two of the most popular open-source tools that have emerged recently are the SVL Simulator (formerly known as LGSVL Simulator) \cite{svl} and CARLA \cite{Dosovitskiy2017}.
These are notable in their use of well-known gaming engines such as Unity Engine \cite{UnityTech} and the Unreal Engine \cite{UE4} respectively. Furthermore, they provide bridges towards open-source Automated Driving System (ADS) such as AutoWare \cite{kato2018} or Apollo \cite{Apollo}. In this paper, we adopt SVL Simulator to virtually test and assess the performance of and Apollo ADS, as required by the IEEE AV Test Challenge 2021.

Virtual simulation platforms enable researchers to develop specific procedures targeting specific components \cite{Pylot,PEM} or to generate synthetic data \cite{Richter2016,talwar2020evaluating}.
A crucial part of the simulation, however, is to design the behaviour of actors, i.e. other road users, such as pedestrians or other vehicles that may be termed Traffic Simulation Vehicles (TSV) or Non-Player Characters (NPC) in SVL.
In fact, while random behaviour could potentially be sufficient, it lacks the structure required for proper testing.
On the other hand, Scenario-based testing \cite{openSC,ScenarioCarla} aims to overcome this by providing a format to describe a scenario in a somewhat deterministic manner, although the AV under test itself may exhibit stochastic behavior.

\section{Virtual Testing Procedure}

The methods we employed for designing the test cases is a mixture of manual and automatic generation, as shown in Figure \ref{fig:pipeline}. In particular, we favoured manual design to promote diversity and completeness, while the role of automation is in providing slight variations of the same test case.

The automated execution of the virtual test cases produces results in both a machine-readable tabular format\footnote{The Virtual Test Results Data Logging format and code is intended to be published at https://github.com/cetran-sg. At the time of publication of this paper, this is pending for review and approval from stakeholders.} (e.g., \texttt{.csv}) which enables \textit{objective} and/or automated evaluation, as well as detailed videos that facilitate offline evaluation on a \textit{subjective} basis, by various domain experts.

We adopt a scenario-based virtual testing approach to test the safety and general driving performance of an autonomous vehicle, which is intended to be deployed in public urban roads under diverse operating conditions. These operating conditions can cover both (a) the pre-specified conditions it is designed to operate in, i.e., its ODD, as well as (b) the actual real-world conditions it must operate in, regardless of whether the ADS was designed to handle those conditions or not, i.e., its OD.

\subsection{Test Categorization}

Our tests have been organized into various types: 
\begin{itemize}
    \item Basic functional/behavioral tests: Focusing on Behavioral Safety of the AV in dynamic urban traffic conditions.
\item Negative tests (edge case tests): Focusing on Safety of the Intended Functionality (SOTIF) and extreme conditions (edge/corner cases) in dynamic urban traffic conditions.
\item Environmental tests: Focusing on the operating conditions (e.g., weather, lighting) around the AV that can change/evolve dynamically.
\item OD/ODD coverage tests: Focusing on special aspects of the OD/ODD in which the AV is designed to operate in. This may also cover out-of-ODD conditions, at which the AV should achieve the minimal risk condition (MRC).
\item Regression tests: Focusing on essential items to test for any regressions made due to changes in the ADS periodically; to be useful in long-term testing as the ADS is being developed. 
\end{itemize}
 
\subsection{Test Objectives}

The tests are designed with two broad level objectives. 
The primary objective is to test the Vehicle under Test (VUT), i.e., the AV, under different categories of conditions. 
Firstly, this should help gain an understanding of its current capabilities in terms of driving safely in a given ODD typically on public roads in an urban environment under different environment conditions including non-optimal weather, traffic and lighting.
Secondly, this should help check whether it meets the general capability expectations as per the rule of the land, to give confidence to the AV developers, regulators and general public, that the AV is safe for public deployment. Such rules may be outlined in the driving regulations for the traffic jurisdiction, e.g., California Driver handbook from DMV or Technical Reference 68 (TR68) in Singapore. 

The second objective is to critically evaluate the usefulness and effectiveness of automated test case generation, in contrast to traditional manual test case development that relies on the skills and knowledge of experienced test personnel, test experts and domain experts. Understanding the strengths and limits of automation helps test engineers to launch a comprehensive test strategy that can make the best use of both modalities.

\subsection{Scenario Diversity}
Specific attention has be dedicated to ensure diversity in testing that can be measured in terms of coverage of various important aspects. In particular, we focus on diversity of Environment (road layout, signals, weather conditions) and Actors, i.e. the other road users that define the traffic scenario.

\subsubsection{Environment}
We primarily choose the San Francisco map available in SVL since it provides a wide variety of road layouts. Additionally, we consider the Borresgas Avenue map for richer detail and higher fidelity of map modeling. 
In fact, one of the first steps is to observe the map and to identify all the relevant road features such as pedestrian crosswalks, traffic light intersections, non-signalized intersection, bus stops, parking areas, and traffic signs. Special road layouts (e.g., skewed or star-shaped intersections) also can be considered and exploited.
Furthermore, our test cases should include a diverse use of traffic lights, weather conditions, and objects on the road such as construction zones or traffic cones that occur in the OD/ODD. Given the scope of the AV Test Challenge, we exploit all the options available in SVL Simulator. These considerations enhances the variety and diversity for our test cases and make them meaningful.

\subsubsection{Actors}
The actual dynamic traffic scenarios are determined by the scripted Actors (or NPCs) that are usually designed to reproduce pre-defined and repeatable behavior and trajectories, and that can be activated (triggered) based on some predefined conditions.
Hence, a comprehensive test case selection should provide a proper distribution across the various actors and objects that can be typically found in the deployment area and/or OD:
\begin{itemize}
    \item Pedestrian: with varying gender, age, size;
    \item Vehicles: cars, trucks, school bus, motorbike, cyclists, emergency vehicles etc.
\end{itemize}

In our framework, we have designed the Actors to perform a variety of maneuvers, e.g., \textit{driving straight}, \textit{turning}, \textit{swerving}, \textit{parking}. 
Furthermore, we also implement advanced scenario-modeling capability such that the Actors can violate traffic rules, to produce additional challenges to the ego-vehicle. This include cases such as a jaywalking pedestrian, or an NPC that may be tailgating the ego vehicle or jumping a red light or simply refusing to give way.

\subsubsection{Known AV weaknesses and Simulation tool limitations}
Known weaknesses in AV technology, particularly, in terms of sensors/perception, prediction, planning and control must be exploited through judicious choice of test parameters.
Furthermore, the limitations of virtual simulation toolchain (e.g., modular testing in SVL directly provides ground truth, offering perfect perception) must be considered and scenario parameters can be adjusted so that the test cases are still effective in finding AV issues.

\section{Scenario-based Testing Framework}
In this section, we describe details of the ViSTA framework which is designed to facilitate virtual validation of a given ADS and simulator, such as Apollo and SVL Simulator.

\subsection{Scenario Generation}
Our abstraction of a Scenario include the below elements:
\begin{itemize}
    \item Map ID: which map the scenario is taking place;
    \item Ego Vehicle Start position and mission:
    determines the ego vehicle starting position and heading, as well as the desired destination coordinates.
    \item a time limit: the simulation timeout, any scenario execution is to be considered a fail if the ego vehicle doesn't reach its destination within the time limit.
\end{itemize}

For specific scenarios, there is the also option to control traffic lights, place traffic cones on the road in specific positions, and specify the weather conditions using the below additional scenario elements.

\begin{itemize}
    \item A list of Actors: the scripted vehicles in the scenario.
    \item A sorted list of time-windows and corresponding weather / lighting condition parameters.
    \item A list of configurations for traffic lights and other controllable objects (e.g., cones).
\end{itemize}

The core distinction between scenarios is the scripted behaviour of the actors and optionally, the applied  dynamic environment conditions (e.g., weather, lighting and/or traffic light conditions). In our approach, the actors and environment have a deterministic behaviour stored in JSON files, which are referred by the relative scenario.

Actor can be of two types, pedestrian and vehicles. Pedestrian are simple to model, as their movement pattern is easily designed by defining the waypoints.
However, vehicles behaviour is more complex, and defining path can be more challenging. 
SVL handling of NPC waypoints is based on linear interpolation which slightly deviates from plausible vehicle dynamics if applied on distant waypoints. However, it is adequate if the waypoints are in close proximity.
Manually designing a scenario is much simpler if its possible to specify only a limited amount of waypoints, and compute the intermediate ones automatically.
Hence, our approach is to facilitates the design with four different but complementary principles.

\paragraph{Key-Waypoint} Our approach is to define key-waypoints, that compose a coarse trajectory.
This trajectory is then refined into a smoother one by a simple adaptation of \cite{quintic}.
In particular, we relaxed the acceleration and jerk limits, since for this task a smooth driving for the scripted vehicles is not crucial. On the contrary, we may actually need to model "bad" drivers with abrupt driving behaviour.
Furthermore, we decided to introduce a speed limit parameter, since we want to have more control in the designing phase.
This way, scripting an Actor is more straightforward, while not giving up a plausible vehicle dynamic. 
\begin{figure*}
    \centering
    \subfloat[b1][Scenario JSON]{\includegraphics[trim=0cm 0.3cm 0cm 0.3cm,width=0.95\columnwidth,clip]{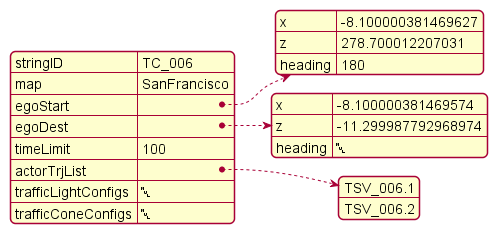}\label{fig:tc_a}}
    \hspace{3mm}
    \subfloat[b1][Actor JSON]{\includegraphics[trim=0cm 0.3cm 0cm 0.3cm,width=0.95\columnwidth,clip]{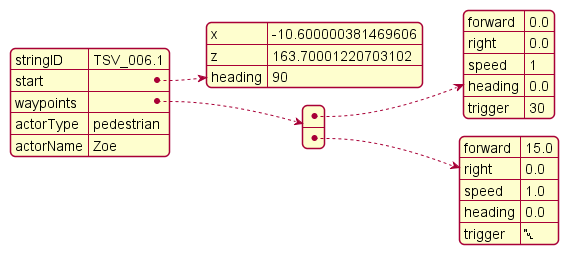}\label{fig:tc_b}}
    \caption{Example of JSON files for TC\_006 and for one the Actor involved.}
    \label{fig:json}
\end{figure*}
\paragraph{Local/Global coordinates} In the design phase, the waypoints are only relative to the previous pose. This allows the designer to not consider global coordinates but rater focus on the vehicle behaviour. By specifying the starting position in the map, all the relative waypoints are easily converted.

\paragraph{Semantic Maneuver} To provide a semantic meaning on the key-waypoints design, we organized the concept of maneuvers in two levels. Level One maneuvers are atomic maneuvers, translate directly into a single key-waypoint, and are parametrized by Level Two maneuvers. The latter are complex maneuvers, that describes a sequence, and a composition, of level one maneuver. 

\paragraph{Parameter Automation} In the design phase, each maneuver requires a set of parameters to be univocally defined. For example, even the simplest "driving straight" can be determined by the length of the segment as well as the target speed. A swerve maneuver, in addition to the previous, is parametrized by the lateral offset induced by the swerving.
Carefully defining each value to generate the appropriate challenge is tedious and time-consuming. In our implementation, it is possible to specify any parameter as a distribution, that will be sampled at the generation phase. Furthermore, we predisposed a function to generate a defined amount of samples for the designed scenario.

Given this functions, during the design phase it is sufficient to specify the sequence of Level Two maneuvers that will be automatically converted, in the order. By definition, a Level Two maneuver is a sequence and combination of Level One maneuvers. Similarly, Level One maneuver is by definition directly converted into a key-waypoint. Finally, the key-waypoint are used to generate the list of intermediate waypoints, i.e. the smooth trajectory. Our approach is to use our adaptation of quintic polynomial planning \cite{quintic}.
The final prepared set of waypoints and corresponding actor/config info can then be stored in a JSON file that can be used runtime to script a deterministic actor (see Figure \ref{fig:json}).

\subsection{Simulation Runtime}
The automatic execution of test cases start by providing a selection of scenario IDs to an automation script that will execute them sequentially and record the results.

The script will establish the necessary connections towards the simulator and Apollo.
A Scenario Manager will read the selected scenario specification (JSON file), and load the map, traffic lights, traffic cones, and weather conditions accordingly.
Furthermore, the script will predispose the Actors and their waypoint instructions, as specified in the associated JSON file.
Then, the script will set the initial state of the ego-vehicle, activate the Apollo modules and provide the target destination.
It will also launch a separate script, running inside docker, to log the ego vehicle and obstacle position obtained on Apollo CyberRT bridge.
After completing these steps, the script will instruct the simulator to run until the ego vehicle's intended mission destination is reached, or the scenario execution time budget is exhausted (times out). Any failure to complete the mission in the allowed time budget, whether it is due to a collision or unsafe situation or not, can lead to the test case flagged as a FAIL, even before it is evaluated offline in further detail (as discussed in next section).

\subsection{Analysis of Virtual Test Execution Results Data and Safety Performance Evaluation}

 When the test cases are executed, the run-time scripts records the dynamic state (position, velocity, heading etc.) of the ego-vehicle and the actors or stationary objects active at each simulation step, into our well-established tabular format as a \texttt{.csv} file.
 These files are then parsed and processed by an offline evaluator scripts that computes various objective metrics to analyze the AV performance. These can include, but not limited to the following.

\paragraph {Occurrence of accidents or collisions}
The ego-vehicle is expected to avoid collisions or accidents with other actors or objects. Exceptions would be when the collision is unavoidable and did not occur due to the ego’s own actions.

\paragraph {Violation of unsafe lateral/longitudinal clearance or safety envelope}
The ego-vehicle is expected to maintain a safety envelope or exclusion zone, represented by lateral/longitudinal clearance distances between ego-vehicle and an actor or object on or beside or relevant to the ego’s path. Exceptions would be when the collision is unavoidable and did not occur due to the ego’s own actions.

\paragraph {Violation of minimal TTC}
Time taken for ego-vehicle to collide with an actor or object in future, at their current velocities and considering their current positions, if they are on a collision course.

\paragraph {Violation of unsafe temporal safety distance}
The time taken to travel the Euclidean distance between the nearest points on the body of the ego-vehicle and that of an actor or object, at the current velocities, irrespective of whether they are on a collision course or not.

\paragraph {Violation of road speed limit}
The ego-vehicle is expected not to exceed the road speed limit of the road it is driving in currently.

 Furthermore, the behaviour of the ego-vehicle in a given scenario and/or context can also be analysed \textit{subjectively}, by the domain experts such as the Chief Tester and supporting Test Analysts (at least 3 people in total). The experts will take into account the objective evaluation and then analyse the results subjectively and make a final decision on the test outcome. If the experts differ in their individual judgement, a consensus or voting can be made between the individual opinions. 
 In many cases, objective judgement may not be possible to achieve in practice; therefore, this is an important tool to be adopted and is flexible and scalable; however, this is dependent on the skillset and experience and knowledge of the experts.
 The following are a few high-level metrics designed to subjectively check the safe behavior of the system under test under different scenario conditions. 

\paragraph {Occurrence of collision [IF]}
Applicable when the Ego vehicle collides with any of the NPCs involved.
\paragraph {Unnecessary swerving [NC]}
Applicable when the Ego vehicle moves forward while changing its heading continuously, without a valid reason. 
\paragraph {Unnecessary braking [NC]}
Applicable when the Ego vehicle slows down, without a valid reason.
\paragraph {Following too close to other road users [NC]}
Applicable when the Ego vehicle is tailgating the leading vehicle or cyclist or similar cases.
\paragraph{Other unacceptable on-road behavioural aspects [IF/NC]}
To be subjectively decided by the experts. 
For continuous improvements of the process, new behavioral aspects can be added as additional metrics as the virtual testing process is executed and the testers gain more experience.

The general evaluation procedure is as follows:
For each test case, we can obtain an evaluation for each metric, which may be an Immediate Failure (IF) or Non-conformity (NC). Immediate Failure would mean that the outcome of the test case is a FAIL. 
In contrast, non-conformity would mean that the outcome is a PASS but with some special conditions. 
Finally, an evaluation with no IF or NC for any metric, would imply a direct PASS for the test case.

The final decision on the outcome of a test case is based on a judicious combination of objective and subjective evaluation, with subjective judgement being final and binding.

\section{Test Cases}
In this section we present a selection of Scenario and Test Cases we developed, summarized in Table \ref{tab:scenario}.
This selection allows to investigate the limitations of Apollo, or SVL, in addressing specific road situations or infrastructures.
\begin{table*}[h]
  \centering
  \caption{Summary of the selected scenarios.}\label{tab:scenario}
    \begin{tabular}{lp{4.3cm}lp{1.2cm}p{3.5cm}p{4.3cm}}
    \textbf{Scenario} & \textbf{Scenario\_description} & \textbf{Test Case} & \textbf{Outcome} & \textbf{Major observations} & \textbf{Apollo problem} \\
    SV\_011 & NPC violates ego's right of way in signalized intersection & TC\_011.a & Pass  & 1. No collision &    -- \\
    SV\_012 & Ego gives way to bus entering bus bay/stop & TC\_012.a & NC    & Ego stopped behind the bus safely; Ego did not managed to overtake the stopped bus & Apollo does not have the capability to detect the bus stop signal in modular testing. Apollo does not overtake \\
    SV\_014 & Ego negotiating construction zone & TC\_014 & Fail  & Ego fails to meet the objective & Apollo does not detect construction zone. in modular testing, there is no ground truth for the objects, therefore Apollo is not detecting. \\
    SV\_016 & Parked truck, pedestrian crossing & TC\_016.d & Pass  & Ego meets objective & Could be due to ground truth due to modular testing (perfect perception) \\    SV\_017 & Ego interacting with other NPC and pedestrians at a complex sequence of intersections & TC\_017.a & Fail  & Ego did not meet the objective;Ego did not able to turn left. & Inconsistent behaviour of Apollo to complete the turn  \\
    SV\_023 & Traffic light signal with arrows  & TC\_023a & Fail  & Ego fails to meet the objective & Ego detects yellow light and still does not move. This is not acceptable based on Singapore rule. \\
    SV\_024 & Traffic signs & TC\_024.a & Pass  & Ego meets objective & Traffic sign not detected in modular test \\
    SV\_027 & Tailgating NPCs & TC\_027.a & Fail  & Ego does not react to tailgating to the NPC & Apollo does not have any evasive maneuvers  \\
    SV\_029 & Accident scenarios & TC\_029.a & Fail  & Ego collided with NPC & Apollo does not predict NPC path when it is out the map. No evasive maneuvers from Apollo \\
    SV\_030 & Weather conditions (time of day) & TC\_30.e & Fail  & Ego too close to the Pedestrian & Since modular testing, no effect \\
    SV\_030 & Weather conditions (worst case) & TC\_30.f & Fail  & Ego too close to the Pedestrian & Since modular testing, no effect \\
    SV\_031 & NPC slows down and stops in front of ego & TC\_031.a & Pass  &       & Inconsistency in overtaking compared to the TC\_012.a \\
    \end{tabular}
\end{table*}

\begin{figure}
    \centering
    \subfloat[a][SV\_016: Modular Test bypass occlusion.]{\includegraphics[trim= 0 0.3cm 0 0.6cm ,width=0.9\columnwidth,clip]{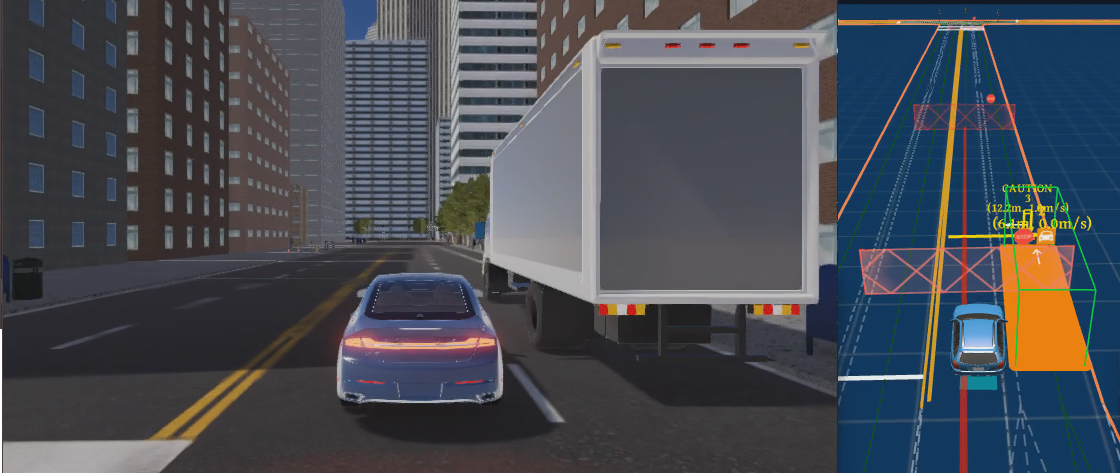}\label{fig:occ}}
    \\
    \subfloat[b][SV\_029: Collision with vehicle exiting parking lot.]{\includegraphics[trim = 0 0.3cm 0 0.6cm ,width=0.9\columnwidth,clip]{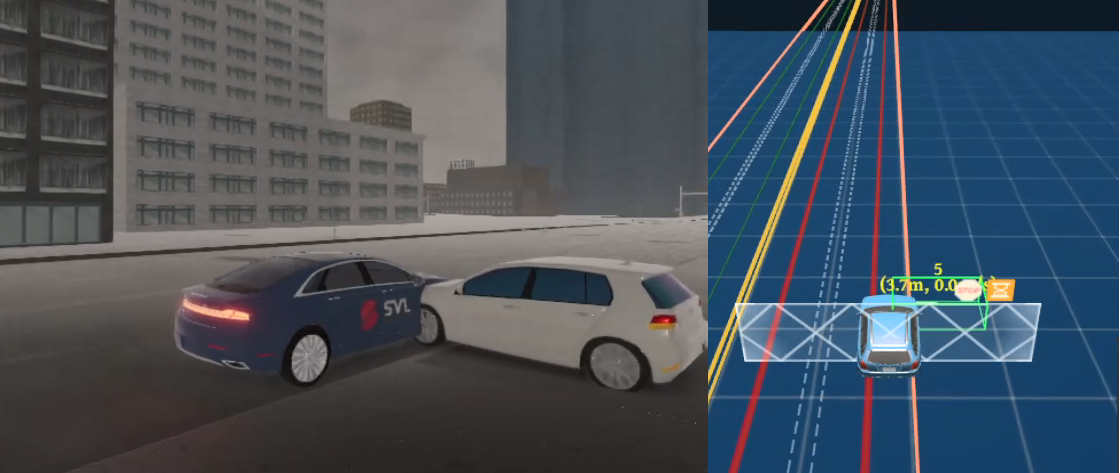}\label{fig:coll}}
    \caption{Example screenshots of 2 test cases during execution.}
    \label{fig:json}
\end{figure}

\paragraph{SV\_011}
In this scenario, the ego-vehicle mission is to cross an intersection. While approaching, an NPC violates its right of way. This would lead to a collision, but the ego seems to predict the imminent collision and reacts by braking. Although this scenario does not require the ego to slow down significantly, its rather fast reaction to the NPC causes the ego to violate traffic rules (e.g., potential to cause rear ending).
\paragraph{SV\_012}
In this scenario, a bus is driving on the left side lane of the ego-vehicle, as both are approaching a bus stop, close to an intersection.
The bus changes lane and stops, the ego reacts accordingly by slowing down and stopping behind the bus. However, the ego-vehicle never attempts to overtake the stationary bus.
This could be proper decision making in presence of a bus stop, but under modular testing, it is not clear whether ego has access to such information.
While there is no unsafe behaviour, it is unclear why there is no overtake attempt, which can potentially cause road hogging.
\paragraph{SV\_014}
The ego-vehicle is approaching a construction zone signalized by traffic cones on the road. Once again, modular testing does not include traffic cones, so that Apollo does not react to them and just drive through.
\paragraph{SV\_016}
In this scenario, a parked truck occludes a jaywalking pedestrian while the ego-vehicle approached.
However, the modular testing configuration of SVL Simulator, which is based on ground truth data, allows the ego-vehicle to detect the pedestrian in all cases (see Figure \ref{fig:occ}). This obfuscates the reason causing the ego-vehicle slowing down, which rather than a cautious approach to an area with low visibility, could be an actual reaction to the perceived danger.

\begin{table*}[h!]
  \caption{Issues we identified using our Test Case Set.}\label{tab:issues}%
    \begin{tabular}{p{0.4cm}p{5.5cm}p{3.8cm}p{4cm}p{0.6cm}p{0.6cm}}
    \textbf{ID} & \textbf{Problem summary} & \textbf{Test conditions} & \textbf{Specific steps to reproduce} & \textbf{SV\_ID} & \textbf{TC\_ID} \\
    1     & Ego slow down a lot when merging & Ego merging between 2 vehicles & Send routing to change lane & SV\_001 & TC\_001 \\
    2    & Inconsistency when merging & Ego merging between 2 vehicles & Send routing to change lane & SV\_001 & TC\_001 \\
    3    & Apollo 5.0 does not have parking capability & Ego is in the driving lane and must park into a gap in between two NPCs. & Send end waypoint position to between 2 parked vehicles & SV\_003 & TC\_003a \\
    4    & Apollo does not perform any evasion maneuver beside slight braking, leading to a near miss & Oncoming vehicle encroaching into Ego Vehicle path & Modular testing enabled & SV\_005 & TC\_005 \\
    5     & Apollo creeps forward when approaching unsignalized intersection even without any crossing vehicle & Ego drives through unsignalized intersection with or without any vehicles &  --   & SV\_001 - 006  & TC\_001 - 006 \\
    6     & Unable to overtake lead vehicle if distance to it is too short & Bus on adjacent lane, cuts in front of ego causing it to brake and stop & Bus suddenly cuts into planned path of ego causing it to jam brakes and unable to overtake Bus & SV\_012 & TC\_012a \\
    7    & Apollo does not have the capability to detect the bus stop in modular testing. & Bus in front of ego & Modular testing enabled & SV\_012 & TC\_012 \\
    8    & Apollo does not detect construction zone; in modular testing, there is no ground truth for the objects, hence Apollo does not react to them & Traffic cones placed in path of ego & Modular testing enabled & SV\_014 & TC\_014 \\
    9     & Unable to change lanes if road too short & short road to change lane before traffic light & Send routing to change lane & SV\_017 & TC\_017a \\
    10    & Inconsistent turning behaviour from Apollo & Ego interacting with other NPCs and pedestrians at a complex sequence of intersections &  --   & SV\_017 & TC\_017a \\
    11    & Apollo takes time to initialize, that could affect the triggering of the test & Ego interacting with other TSVs and pedestrians at a complex sequence of intersections &   --  & SV\_018 & TC\_018 \\
    12    & Map issue. Ego unable to make a u-turn &  --   &  --   & SV\_020 & TC\_020 \\
    13    & Apollo does not have the parking capability. The HD-map does not have the parking lots. & Ego tries to park into parking lot off the map &  --   & SV\_022 & TC\_022 \\
    14    & Ego detects yellow light and still does not move. This is not acceptable based on Singapore rule. &  --   &  --   & SV\_023 & TC\_023 \\
    15    & Traffic sign not detected in modular test & No left turn sign at Market St and Kearny St intersection & Modular testing enabled & SV\_024 & TC\_024 \\
    16    & Apollo cannot do U-turns while routing &  --   &  --   & SV\_026 & TC\_026 \\
    17    & AV cannot respond or give way to tailgating vehicles &  --   &  --   & SV\_027 & TC\_027 \\
    18    & Apollo performs no evasive maneuvers  &  --   &  --   & SV\_027 & TC\_027a \\
    19     & AV cannot detect Emergency Vehicle, or respond or give way &  --   &  --   & SV\_028 & TC\_028 \\
    20& Apollo does not consider NPCs out of the HD-map. No evasive maneuvers from Apollo & NPC exits from carpark outside of map &  --   & SV\_029 & TC\_029 \\
    21     & AV follows a moving pedestrian on road even if adjacent lanes are available. &  --   &  --   & SV\_030 & TC\_30 \\
    22     & AV overtakes a stationary pedestrian on road, dangerously close, even if adjacent lanes are available. &  --   &  --   & SV\_030 & TC\_30 \\
    23    & \mbox{AV swerves left / right due to control latency issues} &  --   &  --   & SV\_031 & TC\_031 \\
    24    & Inconsistency in overtaking &  --   &  --   & SV\_031 & TC\_031 \\
    25     & Unable to avoid/move away from encroaching vehicle & Oncoming vehicle encroaching into Ego Vehicle path & Run autogenerated encroaching scenario & SV\_101 & TC\_101a \\
    \end{tabular}
\end{table*}%

\paragraph{SV\_017}
This scenario takes place at a sequence of skewed intersection, where the the ego vehicle need to turn left after a right turn.
This cause the ego vehicle to be on the wrong lane, with not enough space to change lane. 
Interestingly, a static obstacle on the wrong lane leads the ego vehicle to overtake it, facilitating the lane change and the left turn.

\paragraph{SV\_023}
Traffic lights with arrow signals are not available. Nevertheless, yellow light is detected (due to modular testing), but the ego-vehicle still does not proceed on its path.
This is not a realistic test because the yellow signal duration is not proper. Yellow light was activated for a very long time and the starting point of the ego-vehicle was near the stop line. 
However, this Test Case can be used to test the behaviour of the ego for yellow light signal and as Unit test for yellow light classification robustness.

\paragraph{SV\_024}
Although a straightforward scenario, Apollo does not detect the traffic signal. Even in modular testing, traffic signals appears to be only determined by the HD-map annotations Apollo uses.
\paragraph{SV\_027}An NPC is tailgating the ego-vehicle, but does not slow down at a signalised intersection, and hits the ego.
While the unsafe behaviour is attributed to the NPC, it appears that ego-vehicle is not reacting to it in any way.
\paragraph{SV\_029}
In this scenario, an NPC exits a parking lot on the right of the approaching ego-vehicle.
This parking lot is not annotated in Apollo HD-map, and appears that the ego-vehicle is ignoring the vehicle by not generating a prediction on its path. 
This leads to not detecting the imminent danger and no reaction from the ego-vehicle, which leads to a collision as illustrated in Figure \ref{fig:coll}).
\paragraph{SV\_030}
In this scenario, a pedestrian is walking longitudinally in the middle of the road and stops after a while.
We used this scenario to test weather conditions, but given the modular testing setups it appears there is no effect at all.
Nevertheless, in this scenario is interesting to observe how the ego-vehicle follows the pedestrian by driving at very low speed, and overtakes only when the pedestrian stops. However, the overtake maneuver is not proper and comes very close to the pedestrian, which is a safety concern.
\paragraph{SV\_031}
In this scenario, a leading NPC is stopping in front of the ego-vehicle. The ego reacts accordingly by slowing down and overtakes the stopped vehicle.
Although this is the expected behaviour, it seems to be inconsistent with TC\_012a.
\begin{figure*}[t]
     \centering
     \includegraphics[trim= 2cm 6.5cm 1cm 7cm, width=\textwidth, clip]{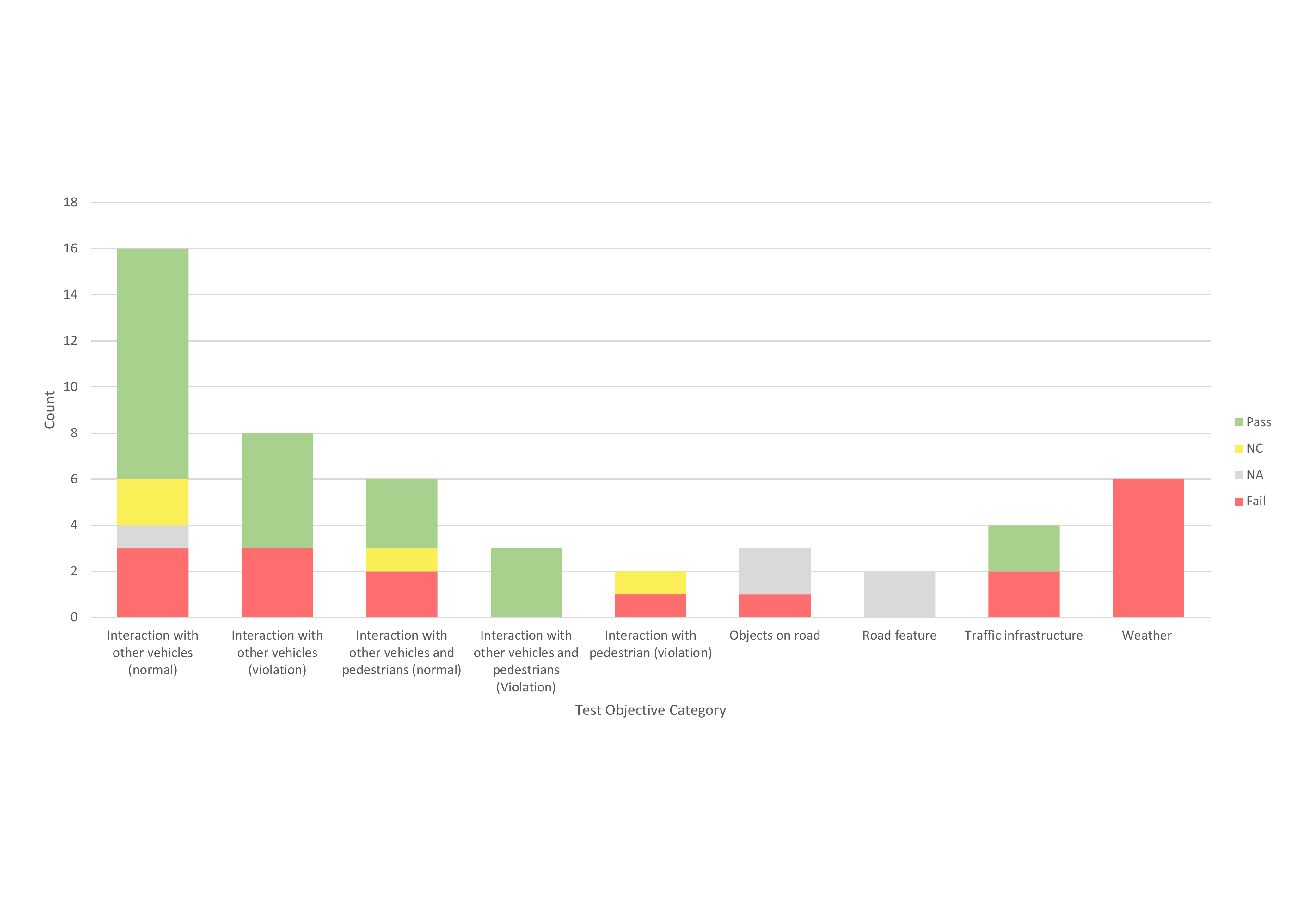}
     \caption{Outcome Distribution of 50 Test Cases organized by the main test objective.}
     \label{fig:result}
\end{figure*}
\subsection{Identified Issues}
In Table \ref{tab:issues} we report all the issues identified using our designed test cases. We can summarize them as follows:
\begin{itemize}
    \item Lack of proper controls: 1, 2, 5, 6, 9, 10, 22, 23, 24. This set contains situations where a better control is desirable, even if it is not necessary to improve safety. 
    \item Lack of functionalities: 3, 13, 16. Apollo 5.0 lacks capability to do Parking and U-turn maneuvers, so we assume that these are  unavailable in the tested configuration.
    \item Lack of evasive maneuver: 4, 17, 18, 21, 25. In this set of scenarios and Test Case appears evident that Apollo is not capable to perform an of evasive maneuver, even when an adjacent empty lane is available.
    \item Implementation issues:  7, 8, 11, 12, 14, 15, 19. These specific Test Cases highlight limitation in the current implementation of the modular testing or SVL-Apollo interaction in general.
    \item Unique situation: 20. This case is particularly interesting since it is very specific. Apparently Apollo does not consider the NPC exiting the parking lot since that area is not part of the HD-map. 
\end{itemize}

\subsection{Summary}
Alternatively, we can classify the Test Cases based on their \textit{main test objective category}, i.e., the most challenging aspect of the scenario that the AV is being tested against in the particular test case.
These challenging aspects can vary based on the AV's interactions with other actors, ranging from normal behaviour to edge cases such as traffic rule violations, to involving specific road features or traffic infrastructure or specific weather conditions.
Figure \ref{fig:result} summarize the outcome of the Test Cases according to the aforementioned classification.
We note that not all of the 50 designed Test Cases were implementable given missing functionalities of the baseline simulator (i.e. no plugins are allowed) or specific road features on the map.
Furthermore, this results are influenced by the modular testing mode, which provides unfair advantages as well as limitations. In particular, test cases involving weather are not effective in modular testing, since any perception challenge is bypassed. Conversely, modular testing also bypass occlusions, which is unrealistic under actual deployment.
Nevertheless, the test case set should also include features that are not directly implementable or testable within the current testing platform or settings. 
\section{Conclusions}
In this paper, we describe details of ViSTA, a virtual testing framework developed to generate scenarios and execute them for validating an ADS driving SAE L4+ AVs.
This enabled us to design specific test cases and identify a wide set of limitations, inconsistencies, and problems within the chosen ADS (Apollo), simulator (SVL), and their integration. 
It is noteworthy that our tests can fail not just in cases when the AV causes accidents or incidents. They can also fail when AV behavior leads to unsafe situations (e.g., TC\_030) that are avoidable through better path planning and responses, such as evasive maneuvers. Also, perfect perception
allows us to focus on such behavioral issues and inconsistencies of the ADS.

This study also highlights the importance of developing better tools for virtual testing, as safety assessment is crucial to ensure the growth and safe deployment of AVs on public roads.
Furthermore, the adoption of interpretable and focused test cases are critical to discover latent issues, that may not necessarily be highlighted when driving in a purely randomized environment.
In future, we hope to extend the ViSTA framework to cover the following aspects. Firstly, we aim to achieve a balanced trade-off between automated and manual design of test cases. Secondly, we hope to implement a robust selection/filtering of meaningful scenario parameters from the large space of feasible parameters for urban driving ODDs, which is a non-trivial task. 
Finally, we plan to include both the actual Sensing and Perception as well as equivalent Perception Error Models \cite{PEM} into the into the virtual testing loop, which can allow further diversity and more effective testing.

\bibliographystyle{./bibliography/IEEEtran}
\bibliography{./bibliography/bib}

\begin{thebibliography}{10}
\providecommand{\url}[1]{#1}
\csname url@samestyle\endcsname
\providecommand{\newblock}{\relax}
\providecommand{\bibinfo}[2]{#2}
\providecommand{\BIBentrySTDinterwordspacing}{\spaceskip=0pt\relax}
\providecommand{\BIBentryALTinterwordstretchfactor}{4}
\providecommand{\BIBentryALTinterwordspacing}{\spaceskip=\fontdimen2\font plus
\BIBentryALTinterwordstretchfactor\fontdimen3\font minus
  \fontdimen4\font\relax}
\providecommand{\BIBforeignlanguage}[2]{{%
\expandafter\ifx\csname l@#1\endcsname\relax
\typeout{** WARNING: IEEEtran.bst: No hyphenation pattern has been}%
\typeout{** loaded for the language `#1'. Using the pattern for}%
\typeout{** the default language instead.}%
\else
\language=\csname l@#1\endcsname
\fi
#2}}
\providecommand{\BIBdecl}{\relax}
\BIBdecl

\bibitem{standard2021j3016}
SAE, ``J3016 standard: Taxonomy and definitions for terms related to driving
  automation systems for on-road motor vehicles,'' 2021.

\bibitem{avchallenge}
``2021 ieee autonomous test driving ai test challenge,''
  \url{http://av-test-challenge.org/}, accessed: 2021-07-23.

\bibitem{Apollo}
``Baidu apollo,'' \url{https://apollo.auto}, accessed: 2021-07-23.

\bibitem{svl}
G.~Rong, B.~H. Shin, H.~Tabatabaee, Q.~Lu, S.~Lemke, M.~Mozeiko, E.~Boise,
  G.~Uhm, M.~Gerow, S.~Mehta, E.~Agafonov, T.~H. Kim, E.~Sterner, K.~Ushiroda,
  M.~Reyes, D.~Zelenkovsky, and S.~Kim, ``{LGSVL} simulator: {A} high fidelity
  simulator for autonomous driving,'' \emph{CoRR}, vol. abs/2005.03778, 2020.

\bibitem{Brockman2016}
G.~Brockman, V.~Cheung, L.~Pettersson, J.~Schneider, J.~Schulman, J.~Tang, and
  W.~Zaremba, ``{OpenAI Gym},'' \emph{arXiv preprint arXiv:1606.01540}, jun
  2016.

\bibitem{Juliani2018}
A.~Juliani, V.-P. Berges, E.~Vckay, Y.~Gao, H.~Henry, M.~Mattar, and D.~Lange,
  ``{Unity: A General Platform for Intelligent Agents},'' \emph{arXiv preprint
  arXiv:1809.02627}, sep 2018.

\bibitem{Dosovitskiy2017}
A.~Dosovitskiy, G.~Ros, F.~Codevilla, A.~L{\'{o}}pez, and V.~Koltun, ``{CARLA:
  An Open Urban Driving Simulator},'' in \emph{Proc. of the 1st Annual
  Conference on Robot Learning}, 2017, pp. 1--16.

\bibitem{UnityTech}
\BIBentryALTinterwordspacing
{Unity Technologies}, ``{Unity Engine},'' 2020. [Online]. Available:
  \url{https://unity.com}
\BIBentrySTDinterwordspacing

\bibitem{UE4}
\BIBentryALTinterwordspacing
{Epic Games}, ``{Unreal Engine 4},'' 2020. [Online]. Available:
  \url{Unrealengine.com}
\BIBentrySTDinterwordspacing

\bibitem{kato2018}
S.~Kato, S.~Tokunaga, Y.~Maruyama, S.~Maeda, M.~Hirabayashi, Y.~Kitsukawa,
  A.~Monrroy, T.~Ando, Y.~Fujii, and T.~Azumi, ``{Autoware on Board: Enabling
  Autonomous Vehicles with Embedded Systems},'' in \emph{Proceedings - 9th
  ACM/IEEE International Conference on Cyber-Physical Systems, ICCPS 2018},
  2018, pp. 287--296.

\bibitem{Pylot}
I.~Gog, S.~Kalra, P.~Schafhalter, M.~A. Wright, J.~E. Gonzalez, and I.~Stoica,
  ``Pylot: {A} modular platform for exploring latency-accuracy tradeoffs in
  autonomous vehicles,'' \emph{CoRR}, vol. abs/2104.07830, 2021.

\bibitem{PEM}
A.~Piazzoni, J.~Cherian, M.~Slavik, and J.~Dauwels, ``Modeling perception
  errors towards robust decision making in autonomous vehicles,'' in
  \emph{Proceedings of the Twenty-Ninth International Joint Conference on
  Artificial Intelligence, {IJCAI-20}}, 2020, pp. 3494--3500.

\bibitem{Richter2016}
S.~R. Richter, V.~Vineet, S.~Roth, and V.~Koltun, ``{Playing for data: Ground
  truth from computer games},'' in \emph{Proc. of the European Conference on
  Computer Vision}, vol. 9906 LNCS.\hskip 1em plus 0.5em minus 0.4em\relax
  Springer Verlag, 2016, pp. 102--118.

\bibitem{talwar2020evaluating}
D.~Talwar, S.~Guruswamy, N.~Ravipati, and M.~Eirinaki, ``Evaluating validity of
  synthetic data in perception tasks for autonomous vehicles,'' in \emph{2020
  IEEE International Conference On Artificial Intelligence Testing
  (AITest)}.\hskip 1em plus 0.5em minus 0.4em\relax IEEE, 2020, pp. 73--80.

\bibitem{openSC}
``Asam open scenario,'' \url{www.asam.net/standards/detail/openscenario/},
  accessed: 2021-07-23.

\bibitem{ScenarioCarla}
``Scenariorunner for carla,'' \url{https://github.com/carla-simulator/},
  accessed: 2021-07-23.

\bibitem{quintic}
A.~Takahashi, T.~Hongo, Y.~Ninomiya, and G.~Sugimoto, ``Local path planning and
  motion control for agv in positioning,'' in \emph{Proc. of the IEEE IROS
  1989}, 1989, pp. 392--397.

\end{thebibliography}

\end{document}